\documentclass[11pt,a4paper]{article}
\usepackage[hyperref]{acl2020}
\usepackage{tikz-dependency}
\usepackage[title]{appendix}
\usepackage{nicefrac}
\usepackage{siunitx}
\usepackage{multirow}
\usepackage{array,booktabs}
\usepackage{times}
\usepackage{amsmath}
\usepackage{dcolumn}
\usepackage{contour}
\usepackage[normalem]{ulem}
\usepackage{graphicx,dblfloatfix,float}

\aclfinalcopy 

\contourlength{0.8pt}

\newcolumntype{d}[1]{D{.}{.}{#1}}

\title{Splitting EUD graphs into trees: A quick and clatty approach}

\author{Mark Anderson \qquad Carlos G\'{o}mez-Rodr\'{i}guez\\
  Universidade da Coru\~na, CITIC \\
  FASTPARSE Lab, LyS Research Group, \\ 
  Departamento de Ciencias de la Computaci\'{o}n y Tecnolog\'{i}as de la Informaci\'{o}n \\
  {\texttt \{m.anderson,carlos.gomez\}@udc.es}}

\date{}

\definecolor{deppink}{HTML}{a92b7a}
\definecolor{depgreen}{HTML}{679033}
\definecolor{deporange}{HTML}{be6320}
\definecolor{depblue}{HTML}{165f77}
\definecolor{depred}{HTML}{b12019}

\definecolor{LightMagenta}{RGB}{249, 249, 255}
\begin{document}

\maketitle
\begin{abstract}
We present the system submission from the FASTPARSE team for the EUD Shared Task at IWPT 2021. We engaged in the task last year by focusing on efficiency. This year we have focused on experimenting with new ideas on a limited time budget. Our system is based on splitting the EUD graph into several trees, based on linguistic criteria. We predict these trees using a sequence-labelling parser and combine them into an EUD graph. The results were relatively poor, although not a total disaster and could probably be improved with some polishing of the system's rough edges.
\end{abstract}

\section{Introduction}

In our group's submission to the IWPT 2020 shared task on EUD parsing \citep{DehAndGomIWPT2020}, we focused on efficiency by applying distillation and training set reduction together with a rule-based approach to convert EUD graphs to UD trees that could be processed by an off-the-shelf parser. Here we describe our entry to the 2021 edition, where we keep the focus on algorithmic simplification of graphs, as well as a prioritisation of efficiency over raw accuracy, but we take the chance to explore different questions that we deem interesting in the context of a breadth-first exploration of the search space of parsing techniques, even if they are not (at least in their current form) competitive in terms of pushing speed or accuracy metrics.

In particular, we wanted to experiment with the application of sequence labelling parsing \citep{strzyz2019viable} to the problem, which we apply to graph parsing for the first time. And more in particular, with the use of a linguistics-oriented approach (à la \citet{DehAndGomIWPT2020}) to guide the parsing process by splitting the EUD graphs into coherent components that can be then parsed by a multitask learning system.


Sequence labelling, the task of assigning one discrete label to each token of a sequence, has long been used for various natural language processing tasks whose output can naturally be represented in this form, such as PoS tagging or named entity recognition. In the case of syntactic parsing, sequence labelling can be applied after defining an encoding that casts each possible syntactic tree for a sentence of length $n$ as a sequence of $n$ labels. While an early attempt to apply it to dependency parsing \citep{spoustov2010dependency} yielded subpar accuracy, the advances in machine learning architectures in the last decade have made this kind of approaches practically viable both for constituency \citep{GomVilEMNLP2018} and dependency parsing \citep{strzyz2019viable}. However, to our knowledge, sequence labelling approaches have not previously been tried for any sort of graph parsing.

\begin{figure*}[b!]
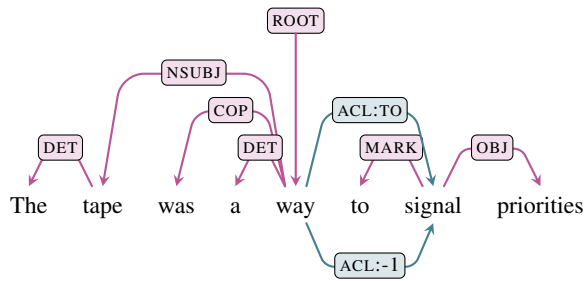

\centering
\begin{dependency}[edge style={deppink!80, thick},label style={fill=deppink!15},edge slant=7]
\begin{deptext}[column sep=0.75em,ampersand replacement=\^,font=\footnotesize]
 The \^ tape \^ was \^ a \^ way \^ to \^ signal \^  priorities \\
\end{deptext}
\depedge{2}{1}{\textsc{det}}
\depedge{5}{2}{\textsc{nsubj}}
\depedge{5}{3}{\textsc{cop}}
\depedge{5}{4}{\textsc{det}}
\depedge{7}{6}{\textsc{mark}}
\depedge[edge style={depblue!80, thick},label style={fill=depblue!15}]{5}{7}{\textsc{acl:to}}
\depedge[edge below, edge height=1.5em, edge slant=7, edge style={depblue!80, thick},label style={fill=depblue!15}]{5}{7}{\textsc{acl:-1}}
\depedge{7}{8}{\textsc{obj}}
\deproot[edge unit distance=3.5ex]{5}{\textsc{root}}
\end{dependency}
\caption{\textbf{Basic} tree split. Only \textsc{acl:to} changes to \textsc{acl:-1} for the relative lemma encoding.}\label{fig:basic}
\end{figure*}

One possible way of extending the search space of a parsing approach is to apply the approach to parse a constant amount of subgraphs (typically, two) whose union provides the final output. This has been applied to go beyond noncrossing dependency trees in transition-based dependency parsing by splitting trees into two subsets of arcs (planes) such that there cannot be crossings within each of them, but their union (the final output) can have crossing arcs \citep{GomNivACL2010,GomNivCL2013,FerGomNAACL2018a}. In semantic parsing, it has also be used to extend the search space from noncrossing graphs to pagenumber-2 graphs by \citet{sun-etal-2017-semantic}, who use graph-based parsing to obtain two noncrossing graphs that are combined by Lagrangian relaxation. In the context of sequence labeling, this approach was recently applied by \citet{strzyz2020bracketing} with similar goals and methods as the transition-based parsers above.

While all these approaches split the output with the goal of relaxing noncrossing constraints, the same can be applied to relax single-head constraints, i.e., go from tree to graph parsing. For example, any graph with in-degree at most 2 can trivially be expressed as the union of two trees. Here we apply that idea in the context of sequence labeling parsing, i.e., we try to generate several sequences of labels (via multitask learning), each of which represents a tree, and which together form an EUD graph.

However, all these splitting approaches share an underlying question: which of the (exponentially many) possible splits is more adequate for the model to adequately learn the parsing problem? The work cited above applies purely algorithmic criteria to choose a canonical split: lazy criteria to minimise the number of plane (subset) switches \cite{GomNivACL2010} or the number of arcs assigned to the second plane \citep{strzyz2020bracketing}, or systematic algorithms that assign crossing arcs to alternating planes \citep{sun-etal-2017-semantic}. These provide splits that are related to the full parse by systemantic structural criteria, but not by linguistic criteria. Since it has been repeatedly shown that it is possible to jointly learn different kinds of dependencies in such a way that they complement each other (e.g. with syntactic and semantic dependencies, as in \cite{HendersonMultilingual,zhou-etal-2020-parsing}) and sequence labeling parsing can benefit from integrating several linguistic representations using multitask learning \citep{StrVilGomACL2019}, what if we try to split parses in a linguistically meaningful way, yielding subsets of dependencies with a distinct meaning that can then be jointly learned? Here we evaluate such an approach.

\section{Splitting graphs}\label{sec:felling_analysis}
\begin{figure}[tbph!]
    \centering
    \includegraphics[width=0.99\linewidth]{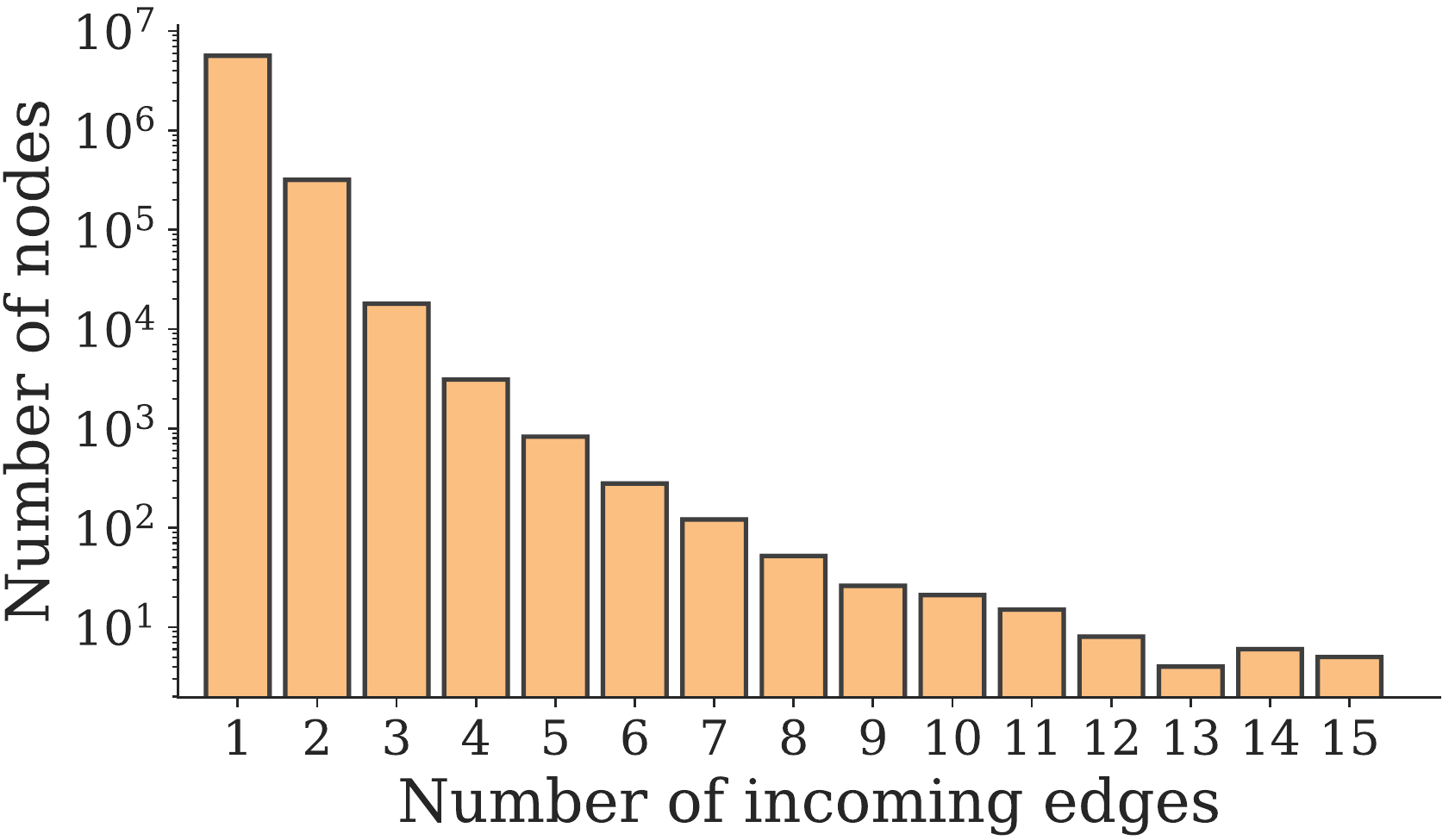}
    \caption{Counts of nodes with x number of incoming edges.}
    \label{fig:incoming_edges_count}
\end{figure}

The vast majority of nodes in a EUD graph only has one incoming edge. If we were to only use one edge per node, we would cover 94.15\% of the edges. Only allowing a maximum of two incoming edges covers 99.53 \% edges, three covers 99.88\%, and four covers 99.95\%. Figure \ref{fig:incoming_edges_count} shows how many nodes have different numbers of incoming edges. Note the logarithmic scale used for the number of nodes. We believe that our graph splitting process results in covering a maximum of two edges, although we have not checked it formally. 
We attempted to split the trees in a linguistically grounded way. We first create what call the basic tree which most closely corresponds to the relative UD tree. We then create a relative, control, and conjunct tree. It is worth noting that, contrary to the work cited in the introduction where parses are split into two disjoint subgraphs, here we have four trees and all these trees can (and usually) overlap. We now describe the different trees into which we split the graph, as well as the collating procedure to combine output trees again into an EUD graph.

\paragraph{Basic tree} This tree is made up of the EUD edges that correspond directly to the UD edges. With one exception for case marking. We add a relative position for the lemma (rather than the lemma itself). This means multi-word case marking is not covered. If no such edge exists we set the edge to (0, root). Although there should be no way this introduces cycles, we check for them anywhere and if any are found we use the Chu-Liu-Edmonds algorithm (CLE) \citep{chuliu65,edmonds1967optimum}, setting scores for expected edges to a sufficiently high value so that they are prioritised, while the others are set very low. If the MST tree has a different edge than in the basic tree, we set that edge to (0, root). If the CLE algorithm changes the ref edge, we change the incoming edge to its head to (0, root). An example is shown in Figure \ref{fig:basic}.

\begin{figure*}[t]
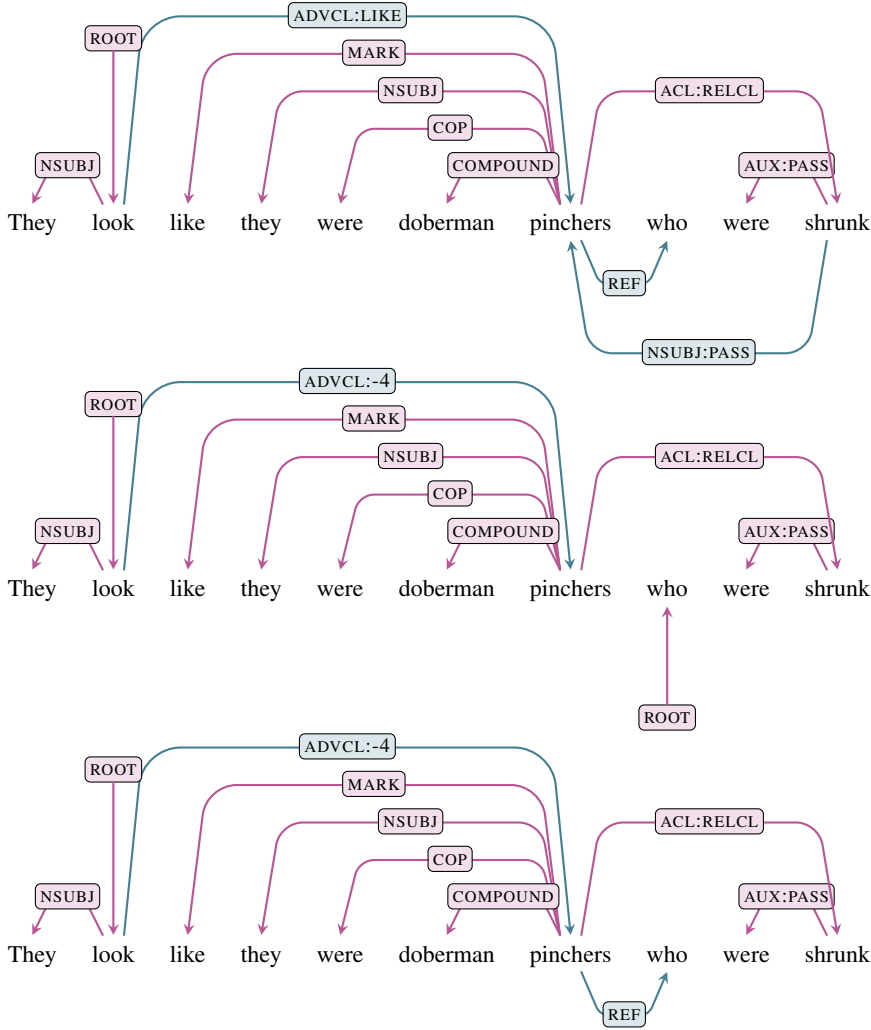

\centering
\begin{dependency}[edge style={deppink!80, thick},label style={fill=deppink!15},edge slant=7]
\begin{deptext}[column sep=0.75em,ampersand replacement=\^,font=\footnotesize]
 They \^ look \^ like \^ they \^ were \^ doberman \^ pinchers \^ who \^ were \^ shrunk\\
\end{deptext}
\depedge{2}{1}{\textsc{nsubj}}
\depedge{7}{3}{\textsc{mark}}
\depedge{7}{4}{\textsc{nsubj}}
\depedge{7}{5}{\textsc{cop}}
\depedge{7}{6}{\textsc{compound}}
\depedge[edge style={depblue!80, thick},label style={fill=depblue!15}]{2}{7}{\textsc{advcl:like}}
\depedge[edge below,edge style={depblue!80, thick},label style={fill=depblue!15}]{10}{7}{\textsc{nsubj:pass}}
\depedge[edge below, edge height=1.5em, edge slant=7, edge style={depblue!80, thick},label style={fill=depblue!15}]{7}{8}{\textsc{ref}}
\depedge{10}{9}{\textsc{aux:pass}}
\depedge{7}{10}{\textsc{acl:relcl}}
\deproot[edge unit distance=3.5ex]{2}{\textsc{root}}
\end{dependency}
\begin{dependency}[edge style={deppink!80, thick},label style={fill=deppink!15},edge slant=7]
\begin{deptext}[column sep=0.75em,ampersand replacement=\^,font=\footnotesize]
 They \^ look \^ like \^ they \^ were \^ doberman \^ pinchers \^ who \^ were \^ shrunk\\
\end{deptext}
\depedge{2}{1}{\textsc{nsubj}}
\depedge{7}{3}{\textsc{mark}}
\depedge{7}{4}{\textsc{nsubj}}
\depedge{7}{5}{\textsc{cop}}
\depedge{7}{6}{\textsc{compound}}
\depedge[edge style={depblue!80, thick},label style={fill=depblue!15}]{2}{7}{\textsc{advcl:-4}}
\depedge{10}{9}{\textsc{aux:pass}}
\depedge{7}{10}{\textsc{acl:relcl}}
\deproot[edge unit distance=3.5ex]{2}{\textsc{root}}
\deproot[edge unit distance=2.5ex,edge below]{8}{\textsc{root}}
\end{dependency}
\begin{dependency}[edge style={deppink!80, thick},label style={fill=deppink!15},edge slant=7]
\begin{deptext}[column sep=0.75em,ampersand replacement=\^,font=\footnotesize]
 They \^ look \^ like \^ they \^ were \^ doberman \^ pinchers \^ who \^ were \^ shrunk\\
\end{deptext}
\depedge{2}{1}{\textsc{nsubj}}
\depedge{7}{3}{\textsc{mark}}
\depedge{7}{4}{\textsc{nsubj}}
\depedge{7}{5}{\textsc{cop}}
\depedge{7}{6}{\textsc{compound}}
\depedge[edge style={depblue!80, thick},label style={fill=depblue!15}]{2}{7}{\textsc{advcl:-4}}
\depedge{10}{9}{\textsc{aux:pass}}
\depedge{7}{10}{\textsc{acl:relcl}}
\depedge[edge below, edge height=1.5em, edge slant=7, edge style={depblue!80, thick},label style={fill=depblue!15}]{7}{8}{\textsc{ref}}
\deproot[edge unit distance=3.5ex]{2}{\textsc{root}}
\end{dependency}
\caption{\textbf{Relative} tree split. Top full EUD graph, middle basic tree, bottom relative tree.}\label{fig:relative}
\end{figure*}

\paragraph{Relative clause tree} We take the basic tree of a graph and replace incoming edges to nodes with \texttt{ref} edges. Again we check for cycles. This tree type was based on an error. We thought that the relative pronoun had two incoming edges: one from the head of the relative clause and one from the referent. This meant we unnecessarily split the basic and relative trees. 
An example of this is shown in Figure \ref{fig:relative}.

\begin{figure*}
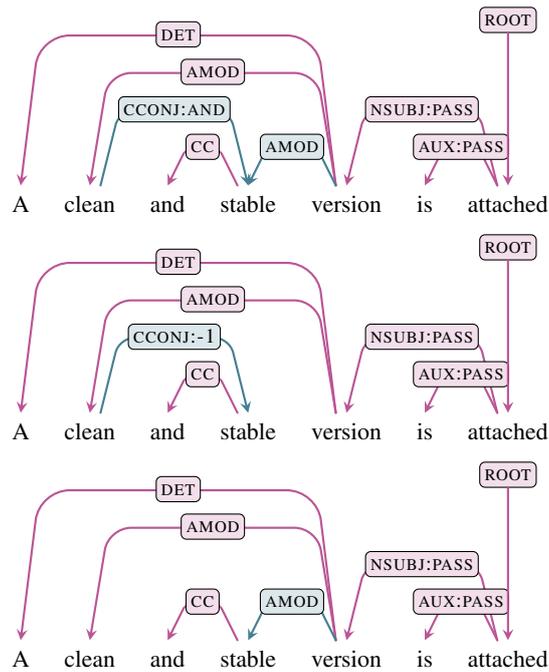

\centering
\begin{dependency}[edge style={deppink!80, thick},label style={fill=deppink!15},edge slant=7]
\begin{deptext}[column sep=0.75em,ampersand replacement=\^,font=\footnotesize]
 A \^ clean \^ and \^ stable \^ version \^ is \^ attached\\
\end{deptext}
\depedge{5}{1}{\textsc{det}}
\depedge{5}{2}{\textsc{amod}}
\depedge{4}{3}{\textsc{cc}}
\depedge[edge style={depblue!80, thick},label style={fill=depblue!15}]{2}{4}{\textsc{cconj:and}}
\depedge[edge style={depblue!80, thick},label style={fill=depblue!15}]{5}{4}{\textsc{amod}}
\depedge{7}{5}{\textsc{nsubj:pass}}
\depedge{7}{6}{\textsc{aux:pass}}
\deproot[edge unit distance=3.5ex]{7}{\textsc{root}}
\end{dependency}

\begin{dependency}[edge style={deppink!80, thick},label style={fill=deppink!15},edge slant=7]
\begin{deptext}[column sep=0.75em,ampersand replacement=\^,font=\footnotesize]
 A \^ clean \^ and \^ stable \^ version \^ is \^ attached\\
\end{deptext}
\depedge{5}{1}{\textsc{det}}
\depedge{5}{2}{\textsc{amod}}
\depedge{4}{3}{\textsc{cc}}
\depedge[edge style={depblue!80, thick},label style={fill=depblue!15}]{2}{4}{\textsc{cconj:-1}}
\depedge{7}{5}{\textsc{nsubj:pass}}
\depedge{7}{6}{\textsc{aux:pass}}
\deproot[edge unit distance=3.5ex]{7}{\textsc{root}}
\end{dependency}

\begin{dependency}[edge style={deppink!80, thick},label style={fill=deppink!15},edge slant=7]
\begin{deptext}[column sep=0.75em,ampersand replacement=\^,font=\footnotesize]
 A \^ clean \^ and \^ stable \^ version \^ is \^ attached\\
\end{deptext}
\depedge{5}{1}{\textsc{det}}
\depedge{5}{2}{\textsc{amod}}
\depedge{4}{3}{\textsc{cc}}
\depedge[edge style={depblue!80, thick},label style={fill=depblue!15}]{5}{4}{\textsc{amod}}
\depedge{7}{5}{\textsc{nsubj:pass}}
\depedge{7}{6}{\textsc{aux:pass}}
\deproot[edge unit distance=3.5ex]{7}{\textsc{root}}
\end{dependency}
\caption{\textbf{Conjunct} tree split. Top full EUD graph, middle basic tree, bottom conjunct tree.}\label{fig:conjunct}
\end{figure*}

\paragraph{Conjunct tree} We start from the basic tree. When an edge is ``conj'' we replace it with the edge in the EUD column that has the same rel as the conj head edge. We use the same cycle check as for the previous trees. An example is shown in Figure \ref{fig:conjunct}.

\begin{figure*}
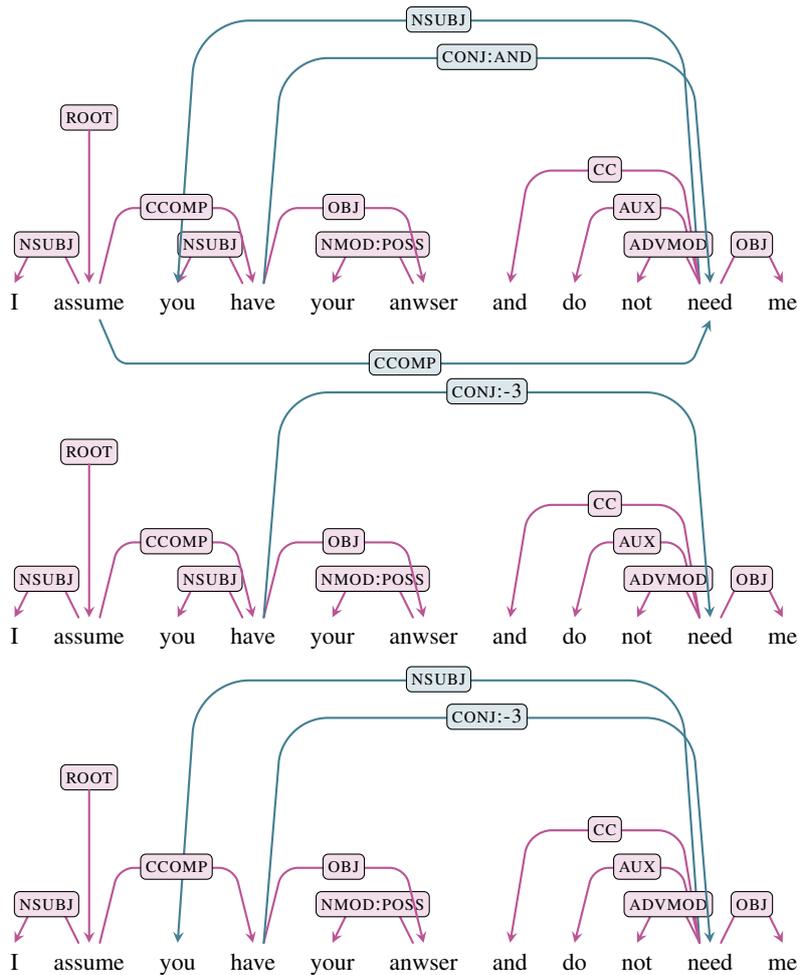

\centering
\begin{dependency}[edge style={deppink!80, thick},label style={fill=deppink!15},edge slant=7]
\begin{deptext}[column sep=0.75em,ampersand replacement=\^,font=\footnotesize]
 I \^ assume \^ you \^ have \^ your \^ anwser \^ and \^ do \^ not \^ need \^ me \\
\end{deptext}
\depedge{2}{1}{\textsc{nsubj}}
\depedge{4}{3}{\textsc{nsubj}}
\depedge[edge style={depblue!80, thick},label style={fill=depblue!15}]{10}{3}{\textsc{nsubj}}
\depedge{2}{4}{\textsc{ccomp}}
\depedge{6}{5}{\textsc{nmod:poss}}
\depedge{4}{6}{\textsc{obj}}
\depedge{10}{7}{\textsc{cc}}
\depedge{10}{8}{\textsc{aux}}
\depedge{10}{9}{\textsc{advmod}}
\depedge[edge style={depblue!80, thick},label style={fill=depblue!15}]{4}{10}{\textsc{conj:and}}
\depedge[edge height=1.5em,edge below, edge style={depblue!80, thick},label style={fill=depblue!15}]{2}{10}{\textsc{ccomp}}
\depedge{10}{11}{\textsc{obj}}
\deproot[edge unit distance=3.5ex]{2}{\textsc{root}}
\end{dependency}
\begin{dependency}[edge style={deppink!80, thick},label style={fill=deppink!15},edge slant=7]
\begin{deptext}[column sep=0.75em,ampersand replacement=\^,font=\footnotesize]
 I \^ assume \^ you \^ have \^ your \^ anwser \^ and \^ do \^ not \^ need \^ me \\
\end{deptext}
\depedge{2}{1}{\textsc{nsubj}}
\depedge{4}{3}{\textsc{nsubj}}
\depedge{2}{4}{\textsc{ccomp}}
\depedge{6}{5}{\textsc{nmod:poss}}
\depedge{4}{6}{\textsc{obj}}
\depedge{10}{7}{\textsc{cc}}
\depedge{10}{8}{\textsc{aux}}
\depedge{10}{9}{\textsc{advmod}}
\depedge[edge style={depblue!80, thick},label style={fill=depblue!15}]{4}{10}{\textsc{conj:-3}}
\depedge{10}{11}{\textsc{obj}}
\deproot[edge unit distance=3.5ex]{2}{\textsc{root}}
\end{dependency}
\begin{dependency}[edge style={deppink!80, thick},label style={fill=deppink!15},edge slant=7]
\begin{deptext}[column sep=0.75em,ampersand replacement=\^,font=\footnotesize]
 I \^ assume \^ you \^ have \^ your \^ anwser \^ and \^ do \^ not \^ need \^ me \\
\end{deptext}
\depedge{2}{1}{\textsc{nsubj}}
\depedge[edge style={depblue!80, thick},label style={fill=depblue!15}]{10}{3}{\textsc{nsubj}}
\depedge{2}{4}{\textsc{ccomp}}
\depedge{6}{5}{\textsc{nmod:poss}}
\depedge{4}{6}{\textsc{obj}}
\depedge{10}{7}{\textsc{cc}}
\depedge{10}{8}{\textsc{aux}}
\depedge{10}{9}{\textsc{advmod}}
\depedge[edge style={depblue!80, thick},label style={fill=depblue!15}]{4}{10}{\textsc{conj:-3}}
\depedge{10}{11}{\textsc{obj}}
\deproot[edge unit distance=3.5ex]{2}{\textsc{root}}
\end{dependency}
\caption{\textbf{Control} tree split. Top full EUD graph, middle basic tree, bottom control tree.}\label{fig:control}
\end{figure*}

 \paragraph{Control tree} We take the basic tree again and this time replace the original nsubj edges of a node when its head has an incoming xcomp or ccomp edge with the other nsubj edge in the EUD graph for the node. We handle potential cycles as usual. An example is shown in Figure \ref{fig:control}. Another error is introduced here, where we don't swap in the \texttt{ccomp} edges.

 \paragraph{Cycles} Only Arabic-PADT has issues with acyclicity after running CLE. So we just collapse the edges that have been changed (this accounts for three instances in the training data).
 \paragraph{Collating trees into graphs} As we operated on a limited time budget, the collating method is egregiously simple. For each node, we take the set of unique edges from all the predicted edges across all trees. When an edge exists between $w_i$ and $w_j$ in more than one tree, we use the label from the first occurrence which is typically that of the basic tree.
 
 Table \ref{tab:gold} shows the EULAS and ELAS when splitting the gold graphs and collating them again. We clearly can cover most of the graph edges with this procedure with Arabic enhanced labels being very low. We believe that this is a bug, but it could be due to some inherent unexpected characteristic of our basic splitting procedure.
 
 \begin{table}[htbp!]
 \small
     \centering
     \begin{tabular}{lp{6em}cc}
     \toprule
     && EULAS & ELAS \\
     \midrule
        ar-padt && 94.04 & 81.91\\
bg-btb && 97.09 & 97.06\\
cs-cac && 94.72 & 93.18\\
cs-fictree && 94.21 & 91.75\\
cs-pdt && 94.41 & 92.36\\
en-ewt && 97.44 & 97.44\\
en-gum && 97.09 & 97.09\\
et-edt && 95.61 & 92.35\\
et-ewt && 95.75 & 91.27\\
fi-tdt && 92.73 & 87.13\\
fr-sequoia && 96.22 & 96.22\\
it-isdt && 96.32 & 95.98\\
lt-alksnis && 94.08 & 87.35\\
lv-lvtb && 93.77 & 93.77\\
nl-alpino && 98.07 & 98.01\\
nl-lassysmall && 97.34 & 97.30\\
pl-lfg && 99.02 & 99.02\\
pl-pdb && 96.37 & 96.19\\
ru-syntagrus && 97.97 & 97.68\\
sk-snk && 96.23 & 94.18\\
sv-talbanken && 96.31 & 96.31\\
ta-ttb && 97.62 & 93.39\\
uk-iu && 96.35 & 95.97\\\bottomrule
     \end{tabular}
     \caption{Graphs formed from splitting the gold annotated development trees and subsequently collating them again.}
     \label{tab:gold}
 \end{table}
 
 \section{Parser}
 We use a BiLSTM network which has word and character embeddings as input. We use a sequence-labelling parser so the edges are predicted as separate labels for each token. Similarly, the edge labels are predicted separetly. But both label predictions are jointly trained with a hard-sharing multi-task architecture with equal weighted loss contributions. UDPipe 2.0 was used for tokenization, lemmatization, and tagging. FastText word embeddings were used but limit vocab space to 50k tokens for memory constraints \cite{bojanowski2017enriching}. We then train 4 parsers for each treebank which are trained on the data generated by splitting the graph, i.e.\ there is one parser for the basic trees, one for the relative trees, and so on. Then the parsers are used to predict their respective tree type and these are all collated to create the predicted graphs for each treebank.
 \paragraph{Sequence labelling parser} (\textsc{SeqLab}) is a parsing approach based on encoding trees as a sequence of one label per token in a sentence, so parsing is reduced to a standard sequence labelling problem \cite{spoustov2010dependency,li-etal-2018-seq2seq,strzyz2019viable}.\footnote{We use refactored encoding/decoding functions from https://github.com/mstrise/dep2label.} We choose to use the original bracketing encoding from \citet{strzyz2019viable}, as it does not require UPOS tags on decoding (the other leading encoding does). While there is a more recent bracketing encoding that covers more non-projectivity \cite{strzyz2020bracketing}, this also involves splitting trees which we assumed would add too much complexity on top of our linguistic-based splitting.
 Our chosen encoding represents a tree as sequence of tags composed of left and right brackets representing each word's incoming and outgoing arcs. Namely, the encoding for $w_i$ is based on:
 \begin{center}
\small
\begin{tabular}{rl}
\\
$<$ ---  & if $\;\epsilon_{j(i-1)}\in\mathcal{E} \wedge j>i-1$  \\[.1cm]
$\setminus$ --- & $\times k \;|$
$k=\sum\limits_{w_j\in S}$
$\begin{cases} 
1 & \text{if}\;\; j<i\wedge\epsilon_{ij}\in\mathcal{E}\\
0 &\text{otherwise} 
\end{cases}$ \\[1.25em]
$/$ --- & $\times k \;|$
$k=\sum\limits_{w_j\in S}$
$\begin{cases} 
1 & \text{if}\;\; i\!-\!1<j\wedge\epsilon_{(i-1)j}\in\mathcal{E}\\
0 &\text{otherwise} 
\end{cases}$\\[1.25em]
$>$ --- & if $\;\epsilon_{ji} \wedge j<i$ 
\end{tabular}
\end{center}

 We repurposed a PyTorch biaffine implementation and edit it to be a simple sequence-labelling system, i.e.\ embedding layers, followed by a number of BiLSTM layers, and one MLP for predicting bracket tags and another for edge labels.  The hyperparameters are shown in Table \ref{tab:experimental_hyperparameters}. The original code for the biaffine is no longer available but a similar version is still available.\footnote{https://github.com/yzhangcs/parser} More details of the system can be found in \citet{anderson2021pareto}. 

\begin{figure}[htbp!]
    \centering
    \begin{dependency}[edge style={depblue!80, thick},label style={fill=depblue!15},edge slant=7]
    \begin{deptext}[column sep=0.4em,ampersand replacement=\^,font=\footnotesize,baseline=(current bounding box.center)]
 That \^ church \^ had \^ a \^ boggin' \^ stench \^ to \^ it \\
 \_ \^ $<\!\setminus$ \^ $<\!\setminus$ \^ $/$
  \^ $<$ \^ $<\!\setminus\setminus\!>$ \^ $/$ \^ $<\setminus\!>$ \\
    \end{deptext}
    \depedge{2}{1}{\textsc{det}}
    \depedge{3}{2}{\textsc{nsubj}}
    \depedge{6}{4}{\textsc{det}}
    \depedge{6}{5}{\textsc{amod}}
    \depedge{3}{6}{\textsc{obj}}
    \depedge{6}{8}{\textsc{nmod}}
    \depedge{8}{7}{\textsc{case}}
    \deproot[edge unit distance=3.ex]{3}{\textsc{root}}
    \end{dependency}%
    \caption{The bracketing encoding introduced by \citet{strzyz2019viable}.}
    \label{fig:sl_example}
\end{figure}
   \begin{figure}[b!]
    \centering
    \includegraphics[width=0.99\linewidth]{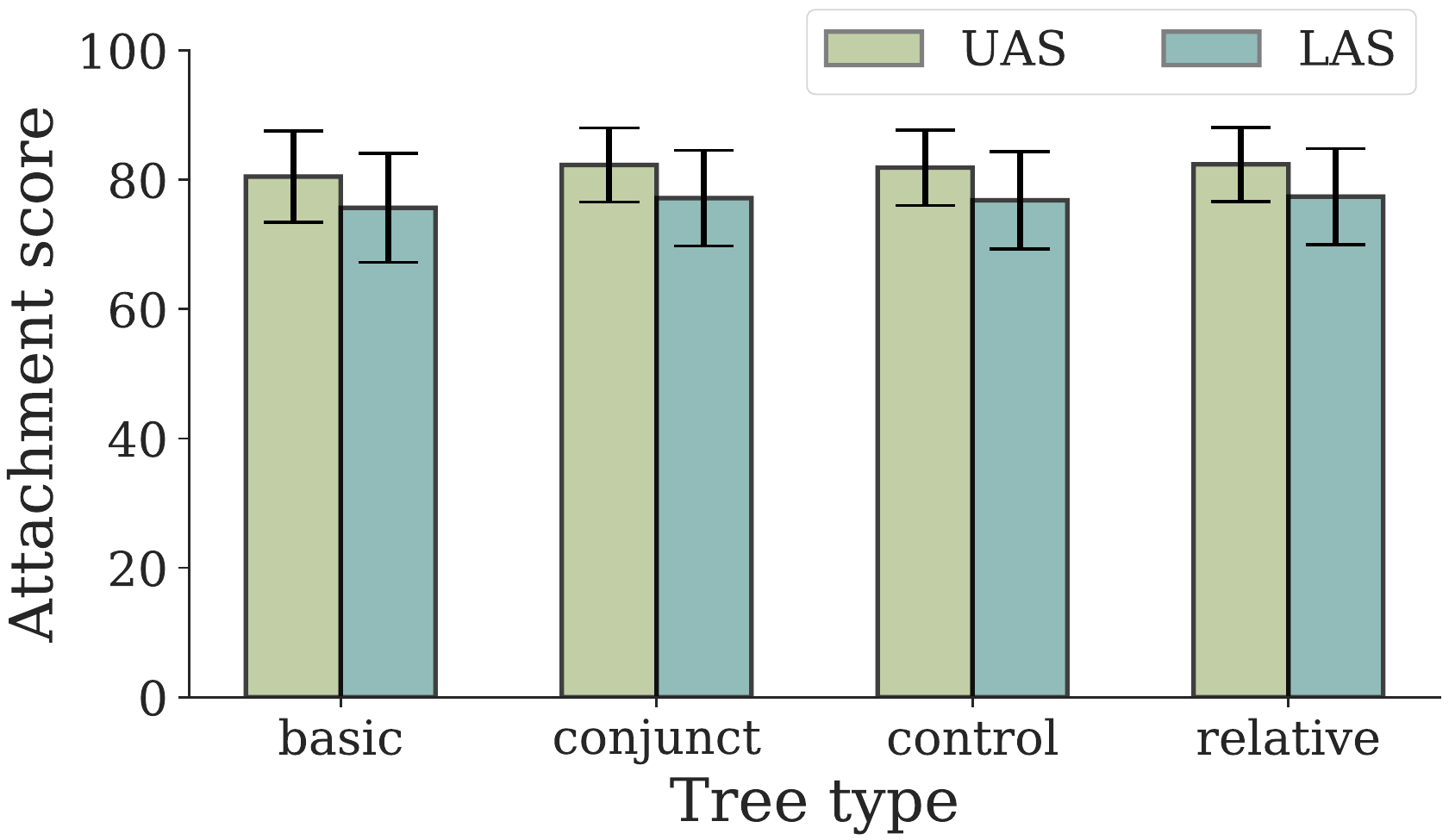}
    \caption{Average performance of parser on each tree type}
    \label{fig:tree_performance}
\end{figure}

\begin{table}[htpb!]
\footnotesize
    \centering
        \tabcolsep=.25cm  
    \begin{tabular}{l p{0.5em} r}
    \toprule
    \textbf{Hyperparameter} & & \textbf{Value}\\
    \toprule
         Word embedding dimensions& & 300\\
         Char embedding dimensions&  & 100\\
         Char BiLSTM dimensions& & 100\\
         Embedding dropout&  & 0.33 \\
         BiLSTM dimensions&  & 500  \\
         BiLSTM layers&  & 3 \\
         BiLSTM dropout&  & 0.33 \\
         MLP layers&  & 1 \\
         Learning rate&  & 0.02 \\
         MLP dropout&  & 0.33 \\
         Momentum&  & 0.9 \\
         L2 norm $\lambda$&  & 0.9\\
         Annealing&  & 0.75$^{\wedge}(\nicefrac{t}{5000})$\\
         $\epsilon$& & 1$\times 10^{-12}$\\
         Optimiser&  & Adam \\
         Loss function&  & cross entropy \\
         Epochs&  & 200 \\
         Min vocab freq. & &1 \\
         Batch size && 32 \\
         Patience && 10 \\
    \bottomrule
    \end{tabular}
        
    \caption{Network hyperparameters.}
    \label{tab:experimental_hyperparameters}
\end{table}

  \begin{figure*}[t!]
    \centering
    \includegraphics[width=0.99\linewidth]{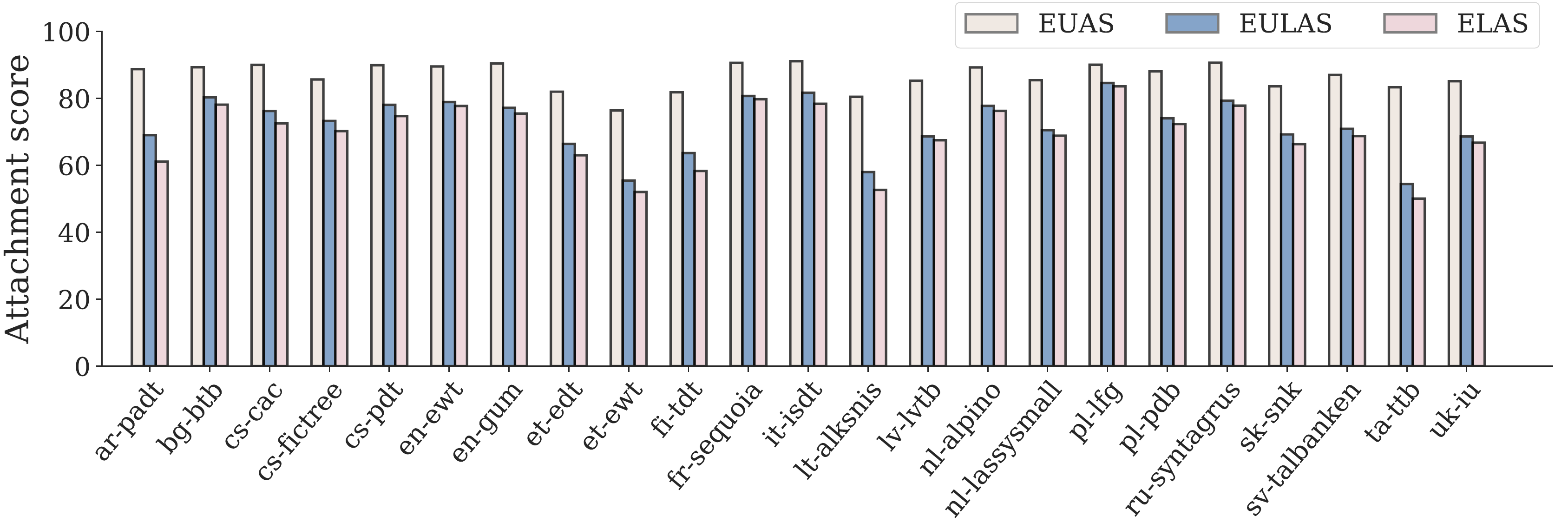}
    \caption{Performance of parser on each treebank.}
    \label{fig:graph_performance}
\end{figure*}
 \section{Results}

The results were rather underwhelming, but our system wasn't an abject failure. Figure \ref{fig:tree_performance} shows the average performance of the parsers trained on each tree type. The performance is pretty stable across each type which is not surprising as the overall structure doesn't vary greatly. But the average performance on the collated trees is quite a bit less as shown in Table \ref{tab:avg_graph_performance}. We decided to include EUAS which measures the unlabelled graph structure. This shows that the parser does learn the graph structure fairly well, but really struggles with labelling the edges. This could be due to appending the labels with the relative positioning of lemmas used for case marking making it harder to predict even the basic label type.
\begin{table}[t!]
    \centering
    \begin{tabular}{lc}
    \toprule
    EUAS & 86.66 \\
   EULAS & 72.02 \\    
    ELAS & 69.21 \\
    \bottomrule
    \end{tabular}
    \caption{Average scores over all treebanks.}
    \label{tab:avg_graph_performance}
\end{table}
Figure \ref{fig:graph_performance} shows the breakdown of the three metrics for each treebank. It is clear that for each treebank a fairly accurate prediction of the graph structure is achieved, but the labelled versions perform much worse. Table \ref{tab:full_results} shows the full results of out system on the test data. The performance across the board is fairly weak and resulted in the worst system which submitted predictions for the full treebank set (and was second last overall).

\begin{table*}[htbp!]
    \centering
    \small
    \tabcolsep=0.041cm
    \begin{tabular}{l*{15}{c}}
    \toprule
        Language&Tokens&Words&Sentences&UPOS&XPOS&UFeats&AllTags&Lemmas&UAS&LAS&CLAS&MLAS&BLEX&EULAS&ELAS\\
        \midrule
Arabic&99.98&94.58&82.09&91.68&88.96&89.14&88.65&90.37&69.84&64.88&59.10&54.18&56.33&61.13&53.74\\
Bulgarian&99.91&99.91&94.17&99.15&97.19&97.95&96.84&97.97&87.85&83.39&78.08&74.62&75.91&80.65&78.73\\
Czech&99.88&99.88&93.18&98.86&95.83&96.01&94.98&98.77&83.75&79.16&74.58&69.89&73.73&76.15&72.85\\
Dutch&99.74&99.74&69.26&96.79&95.29&96.44&94.61&97.06&79.85&74.37&65.23&60.10&63.06&70.49&68.89\\
English&98.38&99.06&88.92&95.85&95.30&94.16&91.39&96.04&82.36&77.99&73.46&64.94&70.55&74.51&73.00\\
Estonian&99.58&99.58&85.60&96.89&97.65&95.78&94.24&94.90&71.70&64.50&59.25&54.75&56.38&63.48&60.05\\
Finnish&99.70&99.68&88.65&97.84&56.14&96.44&54.29&92.11&69.06&62.46&55.58&51.73&51.40&63.20&57.71\\
French&99.65&99.23&94.35&97.05&99.23&91.11&90.28&97.45&84.03&77.14&67.40&57.56&65.57&74.65&73.18\\
Italian&99.93&99.84&98.76&98.52&98.44&98.23&97.66&98.66&88.16&84.92&77.15&73.97&75.79&82.11&78.32\\
Latvian&99.33&99.33&98.74&96.28&89.64&93.79&88.84&95.81&78.37&72.03&67.81&60.47&65.26&67.62&66.43\\
Lithuanian&99.91&99.91&87.87&95.97&90.37&91.07&89.41&93.61&61.39&53.55&47.68&41.70&44.66&52.52&48.27\\
Polish&99.40&99.83&97.52&98.50&93.04&90.80&87.70&97.87&84.32&78.28&73.23&63.04&71.69&74.62&71.52\\
Russian&99.60&99.60&98.80&98.86&99.60&88.97&88.76&98.33&87.09&83.23&79.62&66.43&78.39&80.13&78.56\\
Slovak&100.00&99.99&85.15&97.67&90.14&93.42&89.19&96.47&78.23&71.71&66.18&59.41&63.54&67.09&64.28\\
Swedish&99.18&99.18&93.54&97.25&95.57&88.82&87.63&93.60&78.88&73.11&68.64&56.05&63.96&69.37&67.26\\
Tamil&99.16&94.26&97.52&84.19&82.67&82.27&75.64&88.95&59.80&47.66&42.60&34.50&40.15&46.18&42.53\\
Ukrainian&99.85&99.81&96.61&97.89&94.22&94.18&93.13&97.39&76.26&70.79&65.07&59.24&63.42&65.41&63.42\\
Average&99.60&99.02&91.22&96.43&91.72&92.86&88.43&95.61&77.70&71.72&65.92&58.98&63.52&68.78&65.81\\
\bottomrule
    \end{tabular}
    \caption{Full results on test data per language.}
    \label{tab:full_results}
\end{table*}

\section{Discussion and conclusion}
The relative tree split was based on a mistake. We should have left the \textsc{ref} edges in the basic tree and added the \textsc{nsubj} label variant to the referent in the relative tree. The way it is implemented, we lose those edges. Despite this error, we can still reconstruct most of the edges in the graphs. Beyond this, we can't capture higher-order edges with this method. We did try using a \textsc{Sweep} tree, to capture certain 3rd degree edges. But it seemed as the parser struggled to make sensible predictions and subsequently time ran out before we could test this thoroughly.
    
The collator is very naive. A major issue is introducing extra dummy root edges due to the nature of the split. Another thing we could have tried would have been to collate edges from trees that are only associated with the specific phenomenon of a given tree (i.e.\ conjunct trees only propagating conjunct edges.)

Also, looking at the difference in performance between EUAS and ELAS, it seems the labelling is bad. And the difference between EULAS and ELAS suggests this isn't just a matter of the case marking messing things up. However, the use of relative positional encoding of the case marking might make it harder to learn the labels. Although, the LAS for each tree type isn't that low. So it could potentially be an issue about the way the trees are collated.  Perhaps a first step would be to separate the relative case marking from the relation labels and treat it as a separate task in the MTL system. 

We have presented a simple technique that can easily be extended (and implemented better) but manages to predict relatively accurate unlabelled graphs. It also isn't an utter failure when considering labelled edges, but it seems curious that the performance drops so much compared to the unlabelled performance.

\section*{Acknowledgments}
This work has received funding from the European Research Council (ERC), under the European Union's Horizon 2020 research and innovation programme (FASTPARSE, grant agreement No 714150), from ERDF/MICINN-AEI (ANSWER-ASAP, TIN2017-85160-C2-1-R), from Xunta de Galicia (ED431C 2020/11), and from Centro de Investigación de Galicia ``CITIC'', funded by Xunta de Galicia and the European Union (ERDF - Galicia 2014-2020 Program), by grant ED431G 2019/01. 
\bibliography{acl2020.bib}
\bibliographystyle{acl_natbib}
\end{document}